\documentclass{article}

\usepackage{arxiv}

\usepackage[utf8]{inputenc} 
\usepackage[T1]{fontenc}    
\usepackage{hyperref}       
\usepackage{url}            
\usepackage{booktabs}       
\usepackage{amsfonts}       
\usepackage{nicefrac}       
\usepackage{microtype}      
\usepackage{lipsum}
\usepackage{graphicx}
\usepackage{comment}
\usepackage{here}
\graphicspath{ {./images/} }

\hypersetup{
    colorlinks=true,
    allcolors=green
}

\usepackage[capitalize]{cleveref}
\crefname{section}{Sec.}{Secs.}
\Crefname{section}{Section}{Sections}
\Crefname{table}{Table}{Tables}
\crefname{table}{Tab.}{Tabs.}

\title{SatSwinMAE: Efficient Autoencoding for Multiscale Time-series Satellite Imagery}

\author{
 Yohei Nakayama \\
  Degas Ltd.\\
  \texttt{} \\
   \And
 Jiawei Su \\
  Degas Ltd.\\
  \texttt{} \\
   \And
 Luis M. Pazos-Outón \\
  Degas Ltd.\\
  \texttt{} \\
}

\begin{document}
\maketitle
\begin{abstract}
Recent advancements in foundation models have significantly impacted various fields, including natural language processing, computer vision, and multi-modal tasks. One area that stands to benefit greatly is Earth observation, where these models can efficiently process large-scale, unlabeled geospatial data. In this work we extend the SwinMAE model to integrate temporal information for satellite time-series data. The architecture employs a hierarchical 3D Masked Autoencoder (MAE) with Video Swin Transformer blocks to effectively capture multi-scale spatio-temporal dependencies in satellite imagery. To enhance transfer learning, we incorporate both encoder and decoder pretrained weigths, along with skip connections to preserve scale-specific information. This forms an architecture similar to SwinUNet with an additional temporal component. Our approach shows significant performance improvements over existing state-of-the-art foundation models for all the evaluated downstream tasks: land cover segmentation, building density prediction, flood mapping, wildfire scar mapping and multi-temporal crop segmentation. Particularly, in the land cover segmentation task of the PhilEO Bench dataset, it outpeforms other geospatial foundation models with a 10.4\% higher accuracy.
\end{abstract}

\section{Introduction}
\label{sec:intro}
Earth observation produces a vast amount of unlabeled data on a daily basis. The difficulty of collecting ground truth data presents challenges for training deep learning models via supervised methods. In this scenario, it is critical to develop self-supervised training methods capable of extracting universal data features from the available unlabeled data, and use them effectively to generalize to multiple downstream tasks \cite{wang2022self}. Common approaches for self-supervised learning in the Earth observation domain include conducting masked image modeling \cite{cong2022satmae, reed2023scale, gao2205convmae, noman2024rethinking, jakubik2023foundation} or contrastive learning \cite{he2020momentum, manas2021seasonal, swope2108representation, tarasiou2022embedding}, similarly to the vision domain. Particularly, the Masked Autoencoders (MAE) technique \cite{he2022masked} has been shown effective to pretrain transformer models on unlabeled data and fine-tune them on a broad array of pretrained tasks such as scene classification, land cover segmentation, cloud gap imputation, super-resolution and others \cite{cong2022satmae, reed2023scale, gao2205convmae, noman2024rethinking, jakubik2023foundation}. 

MAE utilizes an asymmetric encoder and decoder architecture based on the Vision Transformer (ViT) \cite{dosovitskiy2020image, he2021mae}. In this approach, masked pixels are reconstructed by the decoder, forcing the model to understand the relationship between pixels. ViT-based models conduct global attention computation on all input tokens, with their computational complexity increasing quadratically with input size. As a consequence, high resolution images must be reduced into manageable sequences via the usage of large patches, losing potentially critical fine-grained information. Furthermore, ViT-based models struggle to recognize objects when there are large variations in the scale of objects in an image.

Swin Transformers \cite{liu2021swin} are proposed for tackling these problems by hierarchically processing the input data. Specifically, the authors propose using a shifted-window mechanism that limits the attention reach to elements inside the window while enabling the exchange of information between separate windows. This reduces computation costs from quadratic to linear with respect to input size. In addition, as features are computed at different scales hierarchically, objects with different sizes can be effectively extracted and passed through the network.

Recently, the Swin Transformer based approaches like SwinMAE \cite{xu2023swinmae} and SwinUNet \cite{cao2022swin} have been proposed for image processing. The Swin Transformer architecture introduces a proximity inductive bias similar to that of CNN models. This allows it to capture both local and global features more effectively. By leveraging the hierarchical attention and shifted-window mechanisms, Swin Transformers reduce the computational demands required by MAE when compared to ViT-based models \cite{chen2021vision}. In particular, SwinUNet uses skip connections to bypass the bottleneck and forward low-level features directly to the decoder. This enables the model to retain rich multi-scale information critical for tasks like segmentation. 

In this work, we extend these approaches to the geospatial domain, where satellite time-series data requires efficient capture of both spatial and temporal dependencies. The Swin architecture’s hierarchical and multiscale capabilities outperform the naive ViT architecture, improving efficiency and handling large-scale, high-resolution datasets. Based on these advantages, we propose a Swin Transformer-based framework for Earth observation tasks.

Our contributions can be summarized as:

\begin{enumerate}
\item We extend the SwinMAE and SwinUNet architectures to support multi-temporal satellite imagery.

\item We pretrain a Swin-based transformer using a 3D MaskedAutoencoder and finetune it on several image-to-image downstream tasks, utilizing skip-connections to preserve multiscale features and a temporal modulation layer to support multi-temporal tasks. 

\item We demonstrate the efficacy of our framework by evaluating our model on five geospatial benchmark datasets, outperforming all state-of-the-art foundation models that were also pretrained solely on images. 

\end{enumerate}

\section{Related work} 
\paragraph{Vision Transformers (ViT) for Image Classification}
Vision Transformers (ViT) have revolutionized the field of image classification by utilizing self-attention mechanisms to capture global relationships within images \cite{dosovitskiy2020image}. Unlike convolutional neural networks (CNNs), which rely on local receptive fields, ViT processes images by dividing them into fixed-size patches and applying self-attention across these patches. This global attention allows ViT to model long-range dependencies, making it highly effective for various vision tasks. However, its computational cost increases significantly with image resolution, posing challenges for high-resolution geospatial data common in remote sensing applications.

\paragraph{Transformer-Based 3D Vision Models}
ViT-based models have been extended to process video data by incorporating the temporal dimension, allowing them to capture both spatial and temporal dependencies \cite{arnab2021vivit}. The have also been applied to geospatial data \cite{pazos2024planted}. However, applying these models to the geospatial domain presents unique challenges. Geospatial datasets often contain a large number of time steps, leading to a significant increase in computational cost due to the quadratic scaling of ViT models with input sequence length. Additionally, the need to process such extensive temporal information places a higher demand on labeled data, which can be scarce in remote sensing, making effective model training even more difficult.

\paragraph{Self-supervised learning in the geospatial domain}  
A defining characteristic of the geospatial domain is the availability of vast unlabeled satellite data. Self-supervised learning has emerged as a powerful approach in this area, enabling the extraction of rich, generalized data representations without the need for manual labeling. This allows models to leverage the abundance of unannotated data, improving their performance on a wide range of downstream tasks.

A common approach in this space is the usage of Masked Autoencoders (MAE) \cite{he2021mae}. Models like SatMAE \cite{cong2022satmae} adapt the MAE framework to effectively handle temporal and multispectral satellite data, showing notable improvements in tasks such as land cover classification and building segmentation. Prithvi \cite{jakubik2023foundation}, another MAE-based model, is pretrained on over 1TB of multispectral data from the Harmonized Landsat-Sentinel dataset, and demonstrates the scalability of MAE for a wide range of geospatial tasks. Prithvi outperforms more specialized, task-specific models in key areas such as flood mapping, burn scar detection, and crop segmentation. Other self-supervised methods, such as contrastive learning (e.g., SeCo \cite{manas2021seasonal}), have also been applied to the geospatial domain, further showcasing the potential of pretraining in this area. 

While these strategies mitigate the challenge of limited labeled data in the geospatial domain, they typically rely on architectures like CNNs or ViT-based models, which face inherent limitations when processing geospatial data. CNNs struggle to capture long-range dependencies due to their localized receptive fields, whereas ViTs, though effective at modeling global attention, become computationally expensive, particularly when dealing with high-resolution satellite imagery common in remote sensing tasks.

\paragraph{Swin Transformers}  
Swin Transformers \cite{liu2021swin} have emerged as a promising alternative to ViTs due to their hierarchical structure and local attention mechanism, which allow them to efficiently process high-resolution images. The Video Swin Transformer extends this approach to the temporal domain, incorporating 3D patch merging and tubelet embeddings to handle spatio-temporal data. MAE has been shown to be an effective approach to pretrain Swin-based models \cite{xu2023swinmae}. In this study, we extend the SwinMAE and SwinUNet architectures along the temporal dimension, following the approach of Video Swin Transformer, to effectively process satellite time-series data. This extension enables the models to capture both spatial and temporal information, which is critical for tasks like change detection and environmental monitoring in remote sensing.

\section{Methodology}
\subsection{Pretraining}
\subsubsection{Dataset} 
For pretraining our model, we use the SSL4EO-S12 dataset \cite{wang2023ssl4eo}, a large-scale earth observation dataset designed for self-supervised learning. This dataset encompasses 250,000 globally distributed locations, each containing multi-seasonal and multimodal satellite imagery from Sentinel missions. Specifically, it includes Sentinel-2 L1C (top-of-atmosphere multispectral), Sentinel-2 L2A (surface reflectance multispectral), and Sentinel-1 GRD (synthetic aperture radar) data, with four images per location, each captured in a different season. To ensure balanced global coverage, the data has been preprocessed, removing Sentinel-2 tiles with more than 10$\%$ cloud cover and eliminating overlapping patches. For our experiments, we use the compressed version of the Sentinel-2 L1C dataset. The dataset is provided in a normalized 8-bit GeoTiff format. Although Sentinel-2 L1C captures 13 spectral bands, we use only six (B2, B3, B4, B8A, B11, B12) during pretraining. To reduce temporal redundancy, we randomly select the first season from each patch and extract data for the subsequent two seasons as input. All image patches are cropped to a resolution of 224×224 pixels, resulting in a final sample size of 3×224×224×6."

\subsubsection{Model Architecture}
\begin{figure*}[t]
  \begin{center}
  \includegraphics[width=\linewidth]{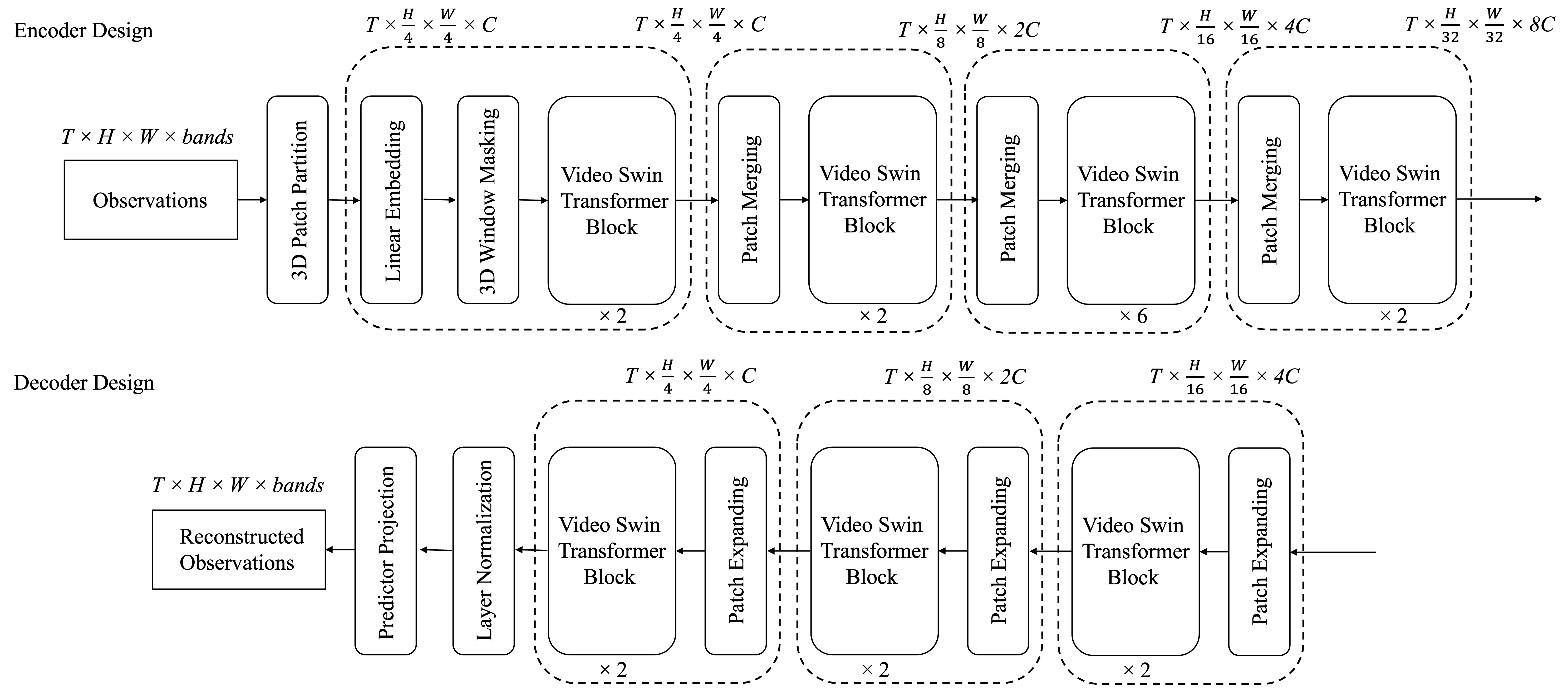}
  \caption{Swin model architecture for MAE pretraining.}
  \label{fig:arc1}
  \end{center}
\end{figure*}

\begin{figure*}[t]
  \centering  
  \includegraphics[width=\linewidth]{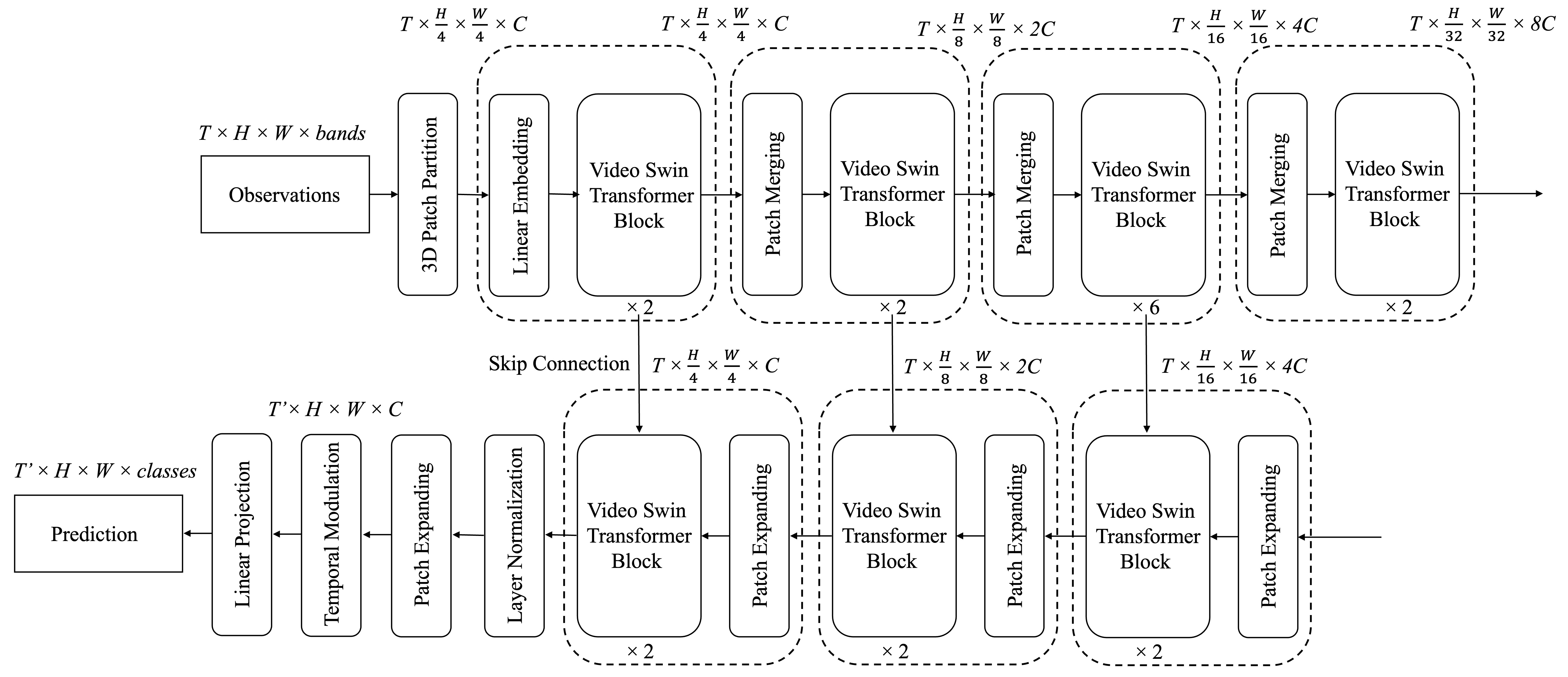}
  \caption{Swin architecture with UNet-like residual connections for finetuning.}
  \label{fig:arc2}
\end{figure*}

Our model architecture follows a similar design as SwinMAE and SwinUNet, with an additional temporal component to handle multi-temporal satellite data, see \cref{fig:arc1}. The model is pretrained using a Masked Autoencoder (MAE) approach, and the pretrained encoder-decoder weights are subsequently fine-tuned for various downstream tasks using the SwinUNet architecture. 

The input consists of satellite observations in the format T×H×W×bands where T is the time dimension, and H and W represent the spatial dimensions. In the encoder, raw inputs are divided into non-overlapping patches through a 3D patch partitioning process,  generating T×H/4×W/4T×H/4×W/4 3D tokens. A linear embedding layer projects each token into a C-dimensional feature space. These Tokens are then passed onto the Swin Transformer blocks, with no patch merging layer applied after the final, similar to the encoder SwinUNet \cite{cao2022swin} used for the downstream tasks. The patch merging layer reduces feature dimensionality by grouping 1×2×2 spatially neighboring patches and applying a linear projection to reduce the feature dimension to half.

For the masking strategy, we adopt a window masking technique \cite{dai2023swin} applied to each spatial slice. This masking strategy randomly selects patches within each slice, akin to the random masking method used in VideoMAE \cite{tong2022videomae}. Instead of removing the masked tokens, they are replaced with a learnable vector, preserving the number of tokens.

In the decoder, Video Swin Transformer blocks and patch-expanding layers are employed. The patch expanding layer performs the reverse operation of patch merging: it first doubles the feature dimension, then spatially expands the resolution of the input features double, and finally reduces the feature dimension to a quarter of its input size. This restores the 3D multi-spectral images to their original shape. Both patch merging and expanding operate only on spatial dimensions, while temporal information remains intact throughout the process.

\subsubsection{Training Settings}
We defined the Swin-Base backbone to have a model size and computational complexity similar to ViT-Base. This architecture can easily scale, by increasing the size of the Swin backbone. 

For all experiments, we set the window size to M = 7 by default, the query dimension of each head to d = 32, and the MLP expansion layer to $\alpha$ = 4. We use the AdamW optimizer with $\beta1 = 0.9$, $\beta2 = 0.999$, batch size of 1536, total epoch of 100, and a one-cycle cosine learning rate scheduler with a maximum learning rate of 1e-5. Pretraining was conducted across 64 NVIDIA A100 GPUs, allowing for efficient large-scale training of the model.

\subsection{Transfer Learning}

In order to adapt our model to perform downstream tasks, we use the Swin-UNet architecture as shown in \Cref{fig:arc2}. The Swin-UNet architecture is similar to the architecture used for MAE pre-training, but introduces several modifications to improve segmentation tasks:
\begin{enumerate}
\item Window masking is removed from the encoder. 

\item Skip connections are across matching resolutions.

\item A patch expanding layer and a temporal modulator are added to the model's head.
\end{enumerate}

Window masking which was used to efficiently capture spatial features during pretraining. This is removed from the encoder in the finetuning stage to allow the model to attend globally to all patches in the input. This enhances the model's downstream performance, particularly for dense prediction tasks, where capturing long-range dependencies without localized constraints is beneficial. Skip connections are introduced between encoder and decoder layers at matching resolutions \cite{cao2022swin}. These connections help mitigate the loss of spatial information caused by downsampling, allowing shallow, fine-grained features to be preserved and combined with deeper, high-level representations. This approach is critical for segmentation tasks where multi-scale contextual information is necessary to predict pixel-level outputs accurately. A patch expanding layer is added to reverse the patch merging process by upscaling the features to their original resolution. This ensures that spatial resolution is maintained when reconstructing the output. Additionally, a temporal modulation layer, composed of a CNN with adjustable kernel sizes, modulates the temporal component of the data. This allows the model to adapt to varying temporal resolutions based on the specific requirements of downstream tasks. Although this framework is primarily designed for image-to-image tasks such as segmentation, the output from the encoder can also be used for classification tasks by adapting the final layers appropriately.

\section{Experiments and Discussion}
\begin{figure*}[t]
  \centering  
  \includegraphics[width=0.7\linewidth]{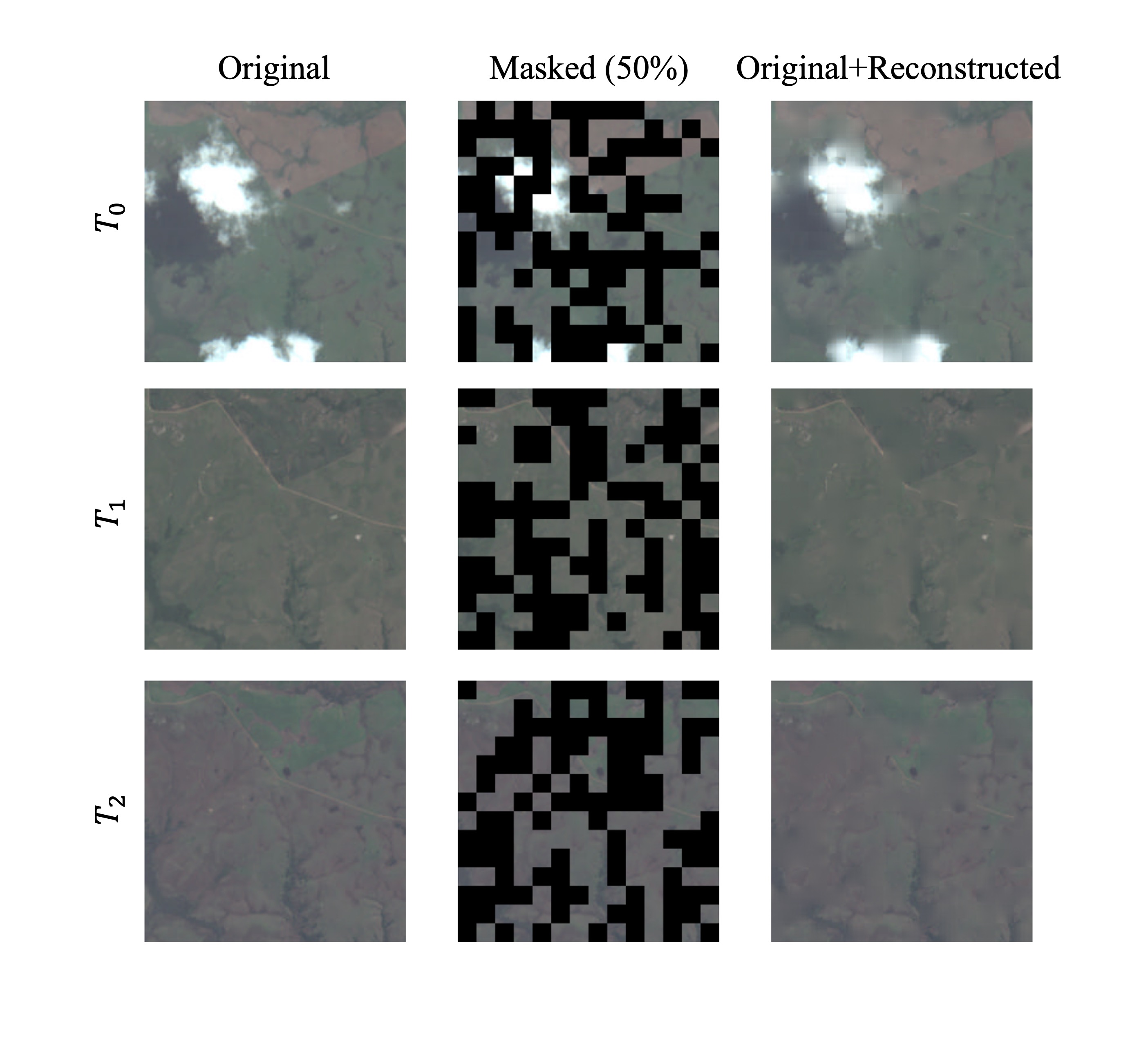}
  \caption{An example of re-constructed satellite imagery during pretraining. Only RGB bands are shown here for visualization purposes.}
  \label{fig:recon}
\end{figure*}
\subsection{Pretraining}
The dataset used for pretraining, SSL4EO-S12 \cite{wang2023ssl4eo}, is notable for its diversity and global coverage, incorporating 250,000 locations with imagery captured across different seasons. By leveraging this multimodal and multitemporal dataset, the model is able to generalize across varied geospatial conditions, which is key to enhancing the model's adaptability.

An example of reconstructed satellite imagery is shown in \Cref{fig:recon}. The figure illustrates the original, masked, and reconstructed satellite images across three seasons, highlighting the robustness of the model in learning spatial and temporal representations. While the RGB bands are displayed for visual clarity, the model actually reconstructs all six multispectral bands used in pretraining (B2, B3, B4, B8A, B11, B12). The success of the reconstruction demonstrates the model’s ability to leverage spatial information across different time steps. Notably, clouds present only in the initial snapshot $T_{0}$ are accurately reconstructed, suggesting that the model effectively combines temporal and spatial information to fill in occluded regions. Pretraining concluded with an MSE loss of 2.65e-3. These pretrained representations will be further evaluated in a range of downstream tasks, including land cover classification and segmentation, as detailed in the following sections.

\subsection{Transfer Learning}
To assess the effectiveness of our pretraining strategy, we evaluate the model's performance on two benchmark frameworks: the PhilEO Bench dataset \cite{fibaek2024phileo}, and the Prithvi Benchmark suite. The former provides three benchmarks; building density estimation,
road segmentation, and land cover classification. The latter includes datasets for flood mapping, wildfire scar mapping, and multi-temporal crop segmentation \cite{jakubik2023foundation}.

\subsubsection{PhilEO Bench Dataset}
PhilEO Bench is a recently introduced global dataset designed to serve as a universal benchmark for geospatial foundation models. It currently it supports 3 different downstream tasks: land cover, buildings and roads. This dataset comprises Sentinel-2 images sampled from geographically diverse regions around the globe, totaling around 400 GB of labeled data for these tasks. In this study, we evaluate our model's performance on the land cover and building density tasks, which involve image-to-image predictions. The land cover task focuses on classifying satellite images into 11 distinct classes based on the ESA World Cover, while the building density task involves predicting the coverage of buildings in squared meters per pixel, ranging from 0 to 100, at a 10m resolution. These tasks test the model's ability to handle different aspects of geospatial data, providing a comprehensive evaluation of our proposed architecture. In both tasks, we evaluate model performance as a function of the available labeled data, demonstrating not only the model's accuracy in data-rich scenarios but also its data efficiency. This evaluation is crucial for remote sensing applications, where labeled data is often scarce, highlighting the model's potential in real-world geospatial tasks.

\begin{figure}
  \centering  
  \includegraphics[width=0.9\linewidth]{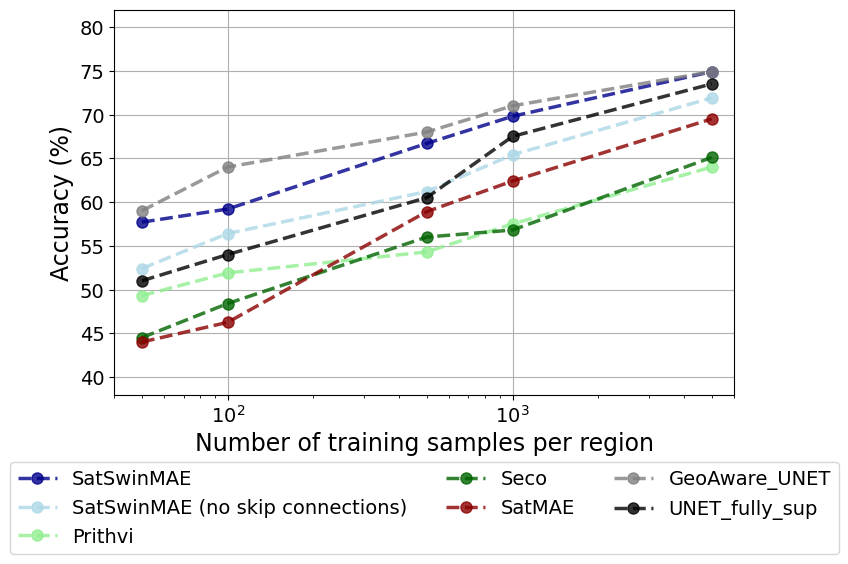}
  \caption{PhilEO Bench land cover classification accuracy.}
  \label{fig:lc acc}
\end{figure}

In \Cref{fig:lc acc}, we compare the land cover segmentation accuracy of our model against the top performing models available in the PhilEO Bench platform. Among others, this includes the FM models (SeCo \cite{manas2021seasonal}, SatMAE \cite{cong2022satmae} and Prithvi \cite{jakubik2023foundation}).

\paragraph{Land Cover}  
In the land cover task, our model achieves better results compared to the existing foundation models (FMs) such as SeCo, SatMAE, and Prithvi. The skip connections applied to our framework lead to an average accuracy improvement of 4.1\%. Notably, the Geo-Aware UNet provided on PhilEO Bench shows similar or slightly better results in this task. Unlike our model, which learns from large-scale unlabeled satellite data in a self-supervised manner, Geo-Aware UNet utilizes a supervised pretraining approach to predict geographical attributes, including coordinates, capture time, and Koppen-Geiger climate zone class, using Sentinel-2 patches.

Despite this advantage, our foundation model without such geo-aware pretraining still outperforms the Geo-Aware UNet in certain scenarios. This suggests that our model's self-supervised learning approach effectively captures pixel-level and temporal patterns inherent in satellite data, even without explicit geographical guidance. We hypothesize that integrating similar geo-aware embeddings or pretext tasks into our training pipeline could further enhance performance, potentially providing the model with a better understanding of spatial-temporal relationships. This represents an exciting direction for future work.
\begin{figure}[b]
  \centering  
  \includegraphics[width=0.9\linewidth]{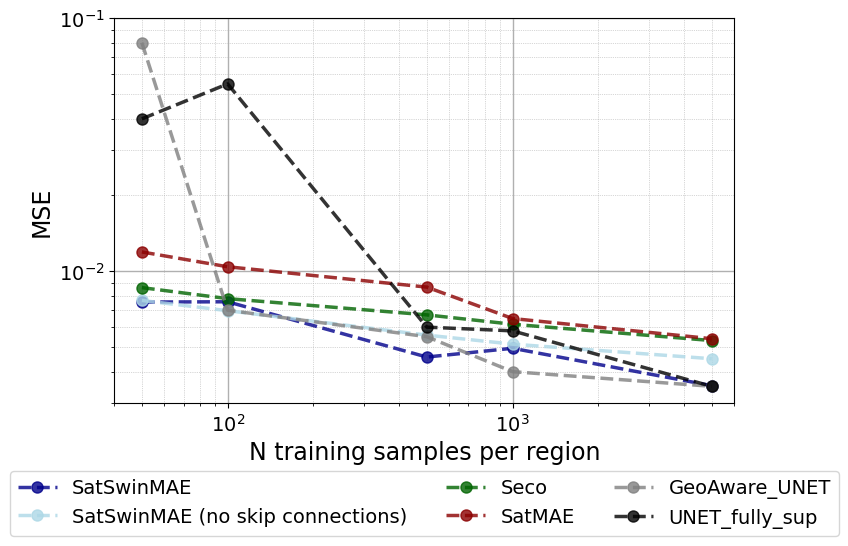}
  \caption{PhilEO Bench building density prediction results.}
  \label{fig:bd acc}
\end{figure}

\paragraph{Building Density}  
In the building density task, our model demonstrates competitive performance against the Geo-Aware model and outperforms all other models trained on similar raw satellite data. For a fair comparison, we report results only for models evaluated at the same 10m resolution. These findings confirm that our approach effectively generalizes across geospatial tasks, handling both classification and regression challenges inherent in satellite imagery analysis.

\subsubsection{Flood Mapping,  Wildfire Scar Mapping and Multi-Temporal Crop Segmentation Dataset}
\begin{table*}
    \centering
    \begin{tabular}{l|ccc}
    \toprule
         Method & IoU & mIoU & mAcc  \\
         & (water) & (both classes) & (both classes)  \\\midrule
         Baseline \cite{bonafilia2020sen1floods11} & 24.21 & --  & --  \\[5px]
         ViT-base \cite{dosovitskiy2020image} & 67.58 & 81.06  & 88.82 \\
         Swin \cite{liu2021swin} & 79.43 & 87.48  & 90.63 \\ 
         Swin (pretrained on ADE20K) \cite{liu2021swin} & 80.58 & 87.98  & 92.02 \\
         
         Prithvi \cite{jakubik2023foundation}  & {82.99} & {90.16} & {94.60}   \\ [5px]
         {SatSwinMAE (no skip connections)}  & {81.08} & {89.24} & {94.76} \\
         \textbf{SatSwinMAE }  & \textbf{84.47} & \textbf{91.12} & \textbf{96.23 } \\
    \bottomrule
    \end{tabular}
    \caption{Sen1Floods11 \cite{bonafilia2020sen1floods11} flood segmentation results.}
    
    \label{tab:flood_mapping}
    \vspace{15px}
    \centering
    \begin{tabular}{l|ccc}
    \toprule
         Method & IoU & mIoU & mAcc  \\
         & (fire scar) & (both classes) & (both classes)  \\\midrule         
         U-Net (DeepLabV3) \cite{chen2017rethinking} & 71.01  & 83.55  & 87.98  \\
         ViT-base \cite{dosovitskiy2020image} & 69.04 & 82.20 & 90.14 \\
         
         {Prithvi \cite{jakubik2023foundation}} & {73.62} & {84.84} & {92.48} \\[5px] 
        {SatSwinMAE (no skip connections)} & {74.37} & {85.63} & {93.21} \\
         \textbf{SatSwinMAE} & \textbf{74.92} & \textbf{85.96} & \textbf{93.44} \\
    \bottomrule
    \end{tabular}
    \caption{Wildfire scars segmentation results.}
    \label{tab:fire_scars}
    \vspace{15px}
    \centering
    \begin{tabular}{l|ccc}
    \toprule
         Method & mIoU & mAcc  \\\midrule         
         U-Net (DeepLabV3) \cite{chen2017rethinking} &  {0.420}& 61.91   \\
         
         {Prithvi \cite{jakubik2023foundation}  }  & {0.426} & {64.06} \\[5px] 
         {SatSwinMAE (no skip connections)} & {0.378}  & {60.31} \\
         \textbf{SatSwinMAE} & \textbf{0.466}  & \textbf{67.68} \\
    \bottomrule
    \end{tabular}
    \caption{Multi-temporal crop type segmentation results.}
    \label{tab:class_performance}
\end{table*}
We evaluated our model on 3 downstream tasks: Flood Mapping,  Wildfire Scar Mapping and Multi-Temporal Crop Segmentation, originally  alongside the Prithvi model \cite{jakubik2023foundation}.

\paragraph{Flood Mapping}  
For flood mapping, we used the Sen1Flood11 dataset, which includes 4,831 chips of size 512×512, covering 14 biomes, 357 ecoregions, and six continents across 11 flood events. 

As shown in \Cref{tab:flood_mapping}, our model achieves the best performance across all metrics: IoU of 84.47, mIoU of 91.12, and mAcc of 96.23. These results demonstrate a significant improvement over Prithvi, which achieved 82.99 in IoU, 90.16 in mIoU, and 94.60 in mAcc. Additionally, compared to the regular Swin Transformer model, our architecture and pretraining approach led to a notable performance increase."

\paragraph{Wildfire Scar Mapping} 
The wildfire scar dataset is created with data from the Monitoring Trends in Burn Severity (MTBS) historical fire database. It comprises 805 scenes collected from 2018 to 2021, with each sample represented by a 512×512 pixel HLS image centered on the wildfire scar.

As presented in \Cref{tab:fire_scars}, our model also achieves state-of-the-art results in this task,  outperforming all other models across all evaluation metrics. This confirms its capability to capture the intricate patterns associated with wildfire scars effectively.

\paragraph{Multi-Temporal Crop Segmentation}  
The multi-temporal crop segmentation task uses Sentinel observations of the Contiguous United States in 2022, paired with crop type labels from the USDA's Crop Data Layer (CDL). Each input comprises three temporal snapshots of Sentinel imagery at a 30 m spatial resolution with a 224 × 224 pixel region.

The results in \Cref{tab:class_performance} show that our outperforms both U-Net and Prithvi, with a mIoU score of 0.466 compared to their scores of 0.420 and 0.426, respectively. This demonstrates a significant improvement, highlighting the effectiveness of our model in handling multi-temporal segmentation tasks.

\section{Conclusion}
In this paper, we proposed an approach to effectively pretrain a foundation model to make it effective across a broad array of key earth observation tasks. To achieve this we leveraged the Swin Transformer's capacity to process multi-scale information. We pretrained the model using an MAE-based approach, with modifications to support the temporal component, and finetuned it using a U-Net inspired network with residual connections to preserve the multi-scale features. We evaluated this model across multiple downstream tasks, including land cover segmentation, building density prediction, flood mapping, wildfire scar mapping, and multi-temporal crop segmentation and achieve a downstream performance comparable or superior to other similar FM models.

\bibliographystyle{unsrt}  
\bibliography{references} 
\end{document}